\documentclass[journal,twoside,web]{ieeecolor}
\usepackage{tmi}
\usepackage{cite}
\usepackage{amsmath,amssymb,amsfonts}
\usepackage{graphicx}
\usepackage{textcomp}
\usepackage{booktabs}
\usepackage{multirow}
\usepackage{url}
\usepackage{bbm}
\usepackage[perpage]{footmisc}
\usepackage{algorithm,algorithmic}
\usepackage{comment}
\usepackage{bm}
\usepackage{dsfont}
\usepackage{mathtools}
\usepackage{adjustbox}
\usepackage{hyperref}

\usepackage{pifont}% http://ctan.org/pkg/pifont

\def\BibTeX{{\rm B\kern-.05em{\sc i\kern-.025em b}\kern-.08em
    T\kern-.1667em\lower.7ex\hbox{E}\kern-.125emX}}
\begin{document}
\title{Compositionally Equivariant Representation Learning}
\author{Xiao Liu, Pedro Sanchez, Spyridon Thermos, Alison Q. O’Neil and Sotirios A. Tsaftaris
\thanks{This work was supported by the University of Edinburgh, the Royal Academy of Engineering and Canon Medical Research Europe by a PhD studentship to Xiao Liu. This work was partially supported by the Alan Turing Institute under the EPSRC grant EP/N510129/1. S.A. Tsaftaris acknowledges the support of Canon Medical and the Royal Academy of Engineering and the Research Chairs and Senior Research Fellowships scheme (grant RCSRF1819\textbackslash8\textbackslash25).}
\thanks{Xiao Liu, Pedro Sanchez and Sotirios A. Tsaftaris are with the University of Edinburgh, Edinburgh, EH9 3FB, UK (e-mail: xiao.liu@ed.ac.uk, pedro.sanchez@ed.ac.uk, s.tsaftaris@ed.ac.uk).}
\thanks{Spyridon Thermos is with Moverse (e-mail: spiros@moverse.ai).}
\thanks{Alison Q. O’Neil is with Canon Medical Research Europe Ltd., Edinburgh, UK and the University of Edinburgh, Edinburgh, EH9 3FB, UK (e-mail: alison.oneil@mre.medical.canon).}
\thanks{Sotirios A. Tsaftaris is also with The Alan Turing Institute, London, UK.}
}

\maketitle
\begin{abstract}
Deep learning models often need sufficient supervision (i.e. labelled data) in order to be trained effectively. By contrast, humans can swiftly learn to identify important anatomy in medical images like MRI and CT scans, with minimal guidance. This recognition capability easily generalises to new images from different medical facilities and to new tasks in different settings. This rapid and generalisable learning ability is largely due to the compositional structure of image patterns in the human brain, which are not well represented in current medical models. In this paper, we study the utilisation of compositionality in learning more interpretable and generalisable representations for medical image segmentation. Overall, we propose that the underlying generative factors that are used to generate the medical images satisfy compositional equivariance property, where each factor is compositional (e.g. corresponds to the structures in human anatomy) and also equivariant to the task. Hence, a good representation that approximates well the ground truth factor has to be compositionally equivariant. By modelling the compositional representations with learnable von-Mises-Fisher (vMF) kernels, we explore how different design and learning biases can be used to enforce the representations to be more compositionally equivariant under un-, weakly-, and semi-supervised settings. Extensive results show that our methods achieve the best performance over several strong baselines on the task of semi-supervised domain-generalised medical image segmentation. Code will be made publicly available upon acceptance at \url{https://github.com/vios-s}. 
\end{abstract}


% \todo{Add Figure 1 explaining compositional equivariance}

\section{Introduction}
\label{sec:introduction}
\label{sec:intro}
When a large amount of labelled training data is available, deep learning techniques have demonstrated remarkable accuracy in medical image analysis tasks like diagnosis and segmentation~\cite{acdc}. However, by contrast, humans are able to learn quickly with only limited supervision, and their recognition is not only fast but also robust and easily generalisable~\cite{tokmakov2019learning, liu2021learning}. For instance, clinical experts tend to remember the compositional components (patterns) of human anatomical structures from medical images they have seen. When searching for anatomy of interest in new images, they use these patterns to locate and identify the anatomy. This compositionality has been shown to enhance the robustness and interpretability in various computer vision tasks~\cite{tokmakov2019learning, huynh2020compositional, kortylewski2020compositional} but has received limited attention in medical applications.

\begin{figure}[t]
\includegraphics[width=0.48\textwidth]{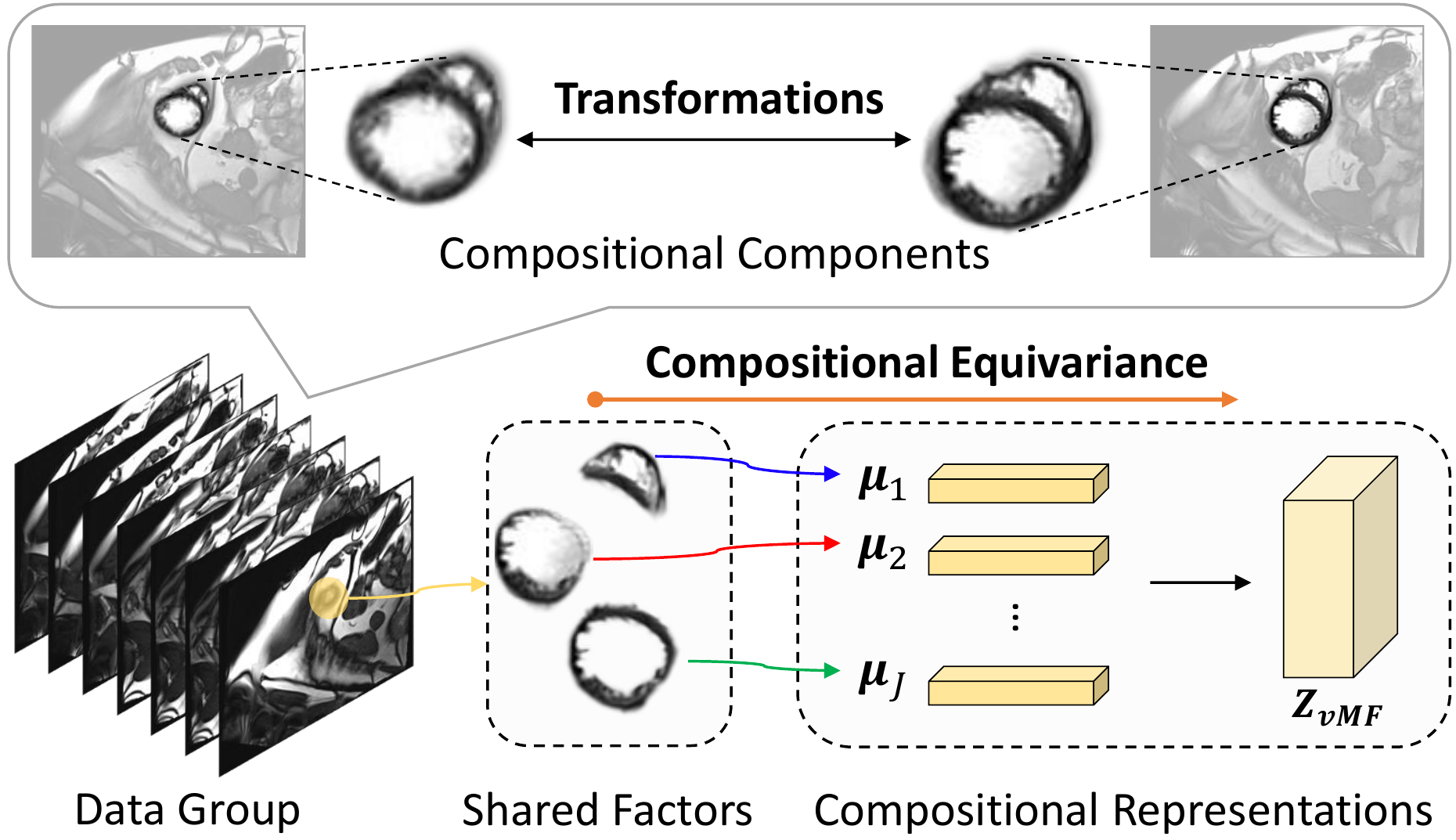}
\centering
\caption{We demonstrate the compositionality and compositional equivariance property. Within a group of data sharing some factors, the compositional components (e.g. the heart) are equivariantly transformed and combined with different components for different images i.e. compositionality. By assigning the information of the factors with the compositional vMF kernels, we can learn compositionally equivariant representations i.e achieving compositional equivariance.} \label{fig::concept}
\end{figure}

Here, we investigate the application of compositionality to learn good representations in the medical field. Drawing inspiration from Compositional Networks~\cite{kortylewski2020compositional}, we model the compositional representations of human anatomy as learnable von-Mises-Fisher (vMF) kernels. Considering that medical images are first processed by deep models into features,  we transform the features into vMF activations that determine the extent to which each kernel is activated at each position. Without any other constraints, the compositional representations do not carry meaningful information that corresponds to the underlying generative factors. We claim that each generative factor is compositional (e.g. the patterns of heart anatomy) and also equivariant for the task, i.e. compositionally equivariant (see Fig.~\ref{fig::concept}). To approximate well the generative factors, we consider different settings i.e. un-, weakly-, and semi-supervised settings and different learning biases that enforce the representations to be more compositionally equivariant. 

% \footnote{vMF kernels are similar to prototypes in~\cite{snell2017prototypical}. However, prototypes are often calculated as the mean of feature vectors for each class using the ground truth masks, while vMF kernels are learned as the cluster centres of the feature vectors.}

To evaluate the level of compositional equivariance, we measure the interpretability and generalisation ability of the representations. We first qualitatively evaluate the interpretability of the activations of each representation for different settings. As expected, we observe that stronger learning biases (e.g. weak supervision or some supervision) lead to better interpretability. 
% Importantly, we find that the learnt compositionally equivariant representations are likely to be used for other tasks without re-training from scratch with extensive annotated data i.e. robust to other tasks. 
Then, we consider the task of semi-supervised domain generalisation~\cite{liu2021semi, liu2022vmfnet, li2018learning} on medical image segmentation and compare our methods with several strong baselines. Extensive quantitative results on the multi-centre, multi-vendor \& multi-disease cardiac image segmentation (M$\&$Ms) dataset~\cite{mnms} and spinal cord gray matter segmentation (SCGM) dataset~\cite{prados2017spinal} show that the compositionally equivariant representations have superior generalisation ability, achieving state-of-the-art performance. 

This work builds on our previously published vMFNet model~\cite{liu2022vmfnet}. Compared to vMFNet: \textbf{a)} we propose compositional equivariance theory; \textbf{b)} we consider more learning settings as well as more design and learning biases to learn the compositional representations. vMFNet is only one out of the five methods; \textbf{c)} moreover, we conduct more experiments, especially on the proposed semi-supervised settings with pseudo supervision and weak supervision on the domain generalisation setting, where better results are observed for some cases compared to vMFNet. We believe that this work demonstrates more comprehensively the benefits and potential of the application of compositionality in the medical domain. In terms of the broader impact of our work, one can easily extend the proposed framework to other equivariant tasks e.g. registration, image translation and multi-model segmentation (see more examples in~\cite{liu2021learning}).

Overall, our \textbf{contributions} are the following:

% \todo{More experiments with weak supervision and see if we emphasize in the contributions.}
% \todo[inline]{@Sotos, I feel the argument that we can evaluate compositional equivariance by evaluating interpretability and generalisation is ambiguous. It is a lack of sufficient support or strict mathematical proof. Do you have any suggestions on dealing with it?}
\begin{itemize}
    \item We revisit the compositionality theory and propose that the generative factors satisfy the compositional equivariance property.
    \item By modelling compositional representations with vMF kernels, we study different settings and different learning biases that can be used to learn compositionally equivariant representations.
    \item We propose a new form of weak supervision i.e. predicting the presence or absence of the anatomical structures.
    \item We evaluate the interpretability and measure the generalisation abilities of the learnt representations as evidence of compositional equivariance. 
    \item We perform extensive experiments on two medical datasets and compare our methods with several strong baselines.
    \item We present extensive qualitative and quantitative results, finding that different learning biases can help to achieve different levels of compositional equivariance. 
\end{itemize}

\section{Related work}

\subsection{Compositionality}
Compositionality has been mostly utilised in robust image classification~\cite{tubiana2017emergence, tokmakov2019learning, kortylewski2020compositional} and recently in compositional image synthesis~\cite{liu2021learning, arad2021compositional}. Among these works, Compositional Networks~\cite{kortylewski2020compositional} --- designed originally for robust classification under object occlusion --- can be adapted to pixel-wise tasks as they learn spatial and interpretable vMF activations. Previous research has combined vMF kernels and activations~\cite{kortylewski2020compositional} for object localisation~\cite{yuan2021robust} and, recently, for nuclei segmentation (with bounding box supervision) in a weakly supervised manner~\cite{zhang2021light}. In this paper, we first model compositional representations using vMF kernels. By incorporating more learning biases that constrain the kernels, we can assign information about each generative factor more specifically to each kernel, resulting in compositional equivariance. Using unlabelled data, we also learn vMF kernels and activations in a semi-supervised manner for domain-generalised medical image segmentation.

\subsection{Domain generalisation}
Various methods have been used to address the domain generalisation problem, such as augmentation of the source domain data~\cite{zhang2020generalising, chen2021cooperative}, regularisation of the feature space~\cite{carlucci2019domain, huang2021fsdr}, alignment of the source domain features or output distributions~\cite{li2020domain}, design of robust network modules~\cite{gu2021domain}, or the use of meta-learning to adapt to possible domain shifts~\cite{dou2019domain, liu2020shape, liu2021feddg, liu2021semi}. Most of these approaches consider the fully supervised learning. More recently, a gradient-based meta-learning model was proposed to handle semi-supervised domain generalisation by integrating disentanglement~\cite{liu2021semi}. Another method used a pre-trained ResNet as a backbone feature extractor, augmenting the source data, and leveraging the unlabelled data through pseudo-labelling~\cite{yao2022enhancing}. Our approach aligns image features to the same von-Mises-Fisher distributions to handle domain shifts. In the semi-supervised setting with reconstruction, the reconstruction further enables the model to handle domain generalisation with unlabelled data. For the semi-supervised setting with weak/pseudo supervision, the weak/pseudo supervision enables the model to be trained with weakly-labelled or unlabelled data.

\section{Method}
\label{sec:method}
In the following, we denote $x$ as a scalar, $\mathbf{x}$ as a vector and $\mathbf{X}$ as a tensor. Consider a dataset $\mathcal{D}=\{\mathbf{X}_{i}, \mathbf{Y}_{i}\}_{i=1}^{N}$ that is defined on a joint space $\mathcal{X} \times \mathcal{Y}$, where $\mathbf{X}_{i}$ is the $i^{th}$ training datum with corresponding ground truth label $\mathbf{Y}_{i}$ (e.g. for a segmentation task, $\mathbf{Y}_{i}$ is the ground truth segmentation mask), and $N$ denotes the number of training samples. We aim to learn a model containing a representation encoding network $\bm{F}_\psi: \mathcal{X} \to \mathcal{Z}$ to extract the representations, and a task network $\bm{T}_\theta: \mathcal{Z} \to \mathcal{Y}$ to perform the downstream task, where $\psi$ and $\theta$ denote the network parameters.

\subsection{Compositionality theory}
Finding good latent representations for the task at hand is fundamental in machine learning~\cite{bengio2013representation,scholkopf2021ieee}. When supervision is available for the latent representations (the ground truth generative factors) and the downstream task (the ground truth labels), it is natural to train $\bm{F}_\psi$ and $\bm{T}_\theta$ with supervised losses as in the Concept Bottleneck Model~\cite{koh2020concept}. However, in practice, usually not all of the generative factors are known. When there is insufficient supervision for either the latent representations or the downstream task, learning generalisable and interpretable representations is a challenging problem to solve. To tackle this issue, we propose to use compositional equivariance as an inductive bias to learn the latent representations. We later show that with the compositional equivariance, it is possible to learn the desired representations without any supervision, with weak supervision, or with some supervision i.e. un-, weakly-, semi-supervised settings.

\begin{figure*}[ht]
\includegraphics[width=0.9\textwidth]{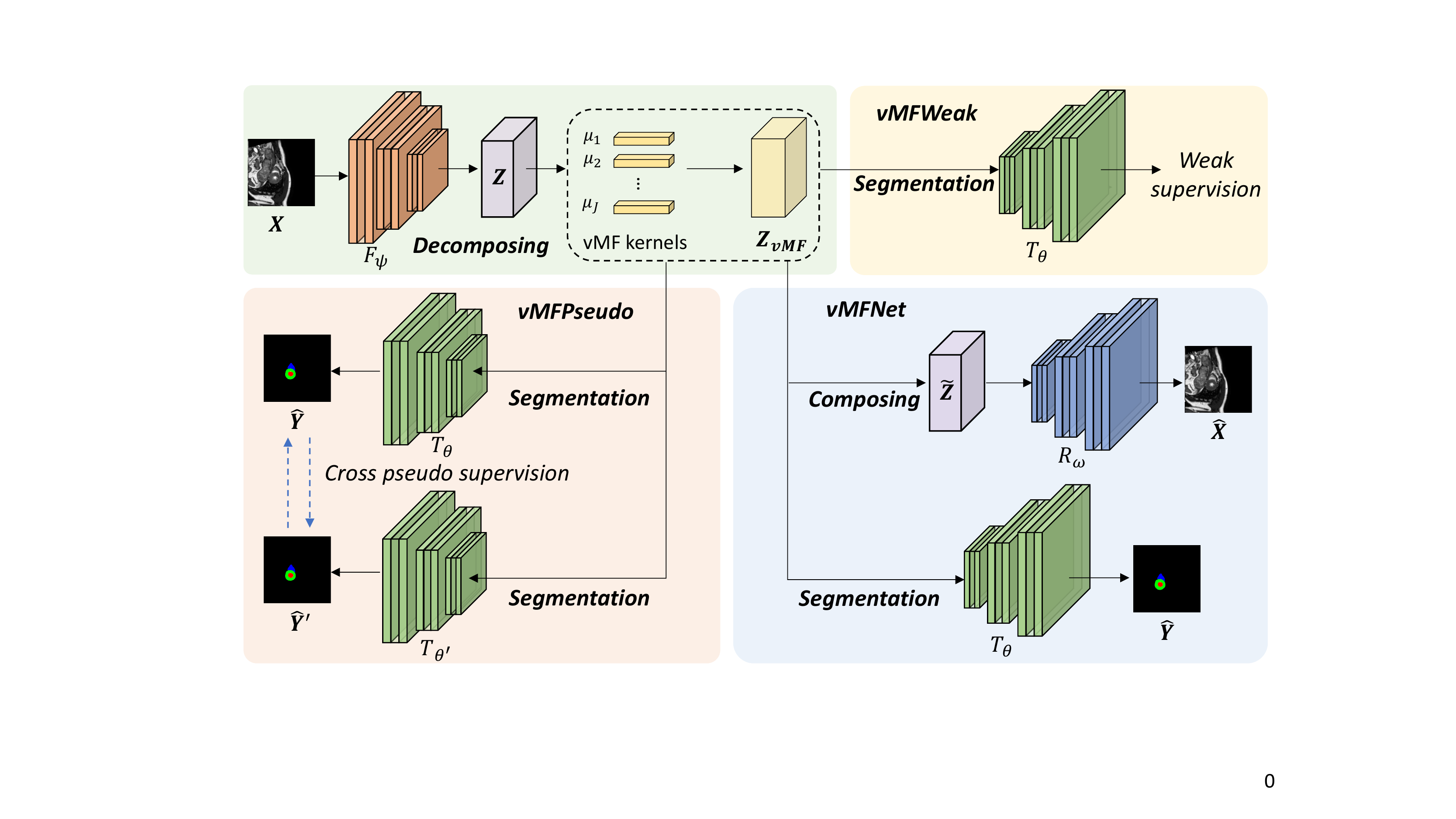}
\centering
\caption{Overall model design for vMFWeak, vMFNet and vMFPseudo. For vMFNet, apart from decomposing and composing modules, the segmentation module is used to predict the segmentation mask by taking the vMF activations as input. For vMFPseudo, we simultaneously train two models and use the prediction of one model as the pseudo supervision for the other model. For vMFWeak, we apply the weak supervision after the output of the segmentation module.} \label{fig::allmodel}
\end{figure*}

% \begin{figure}[t]
% \includegraphics[width=0.4\textwidth]{Decomp.pdf}
% \centering
% \caption{The decomposing module. $\mathbf{Z}$ is the features encoded by a feature encoder network. The feature vector $\mathbf{z}_i\in \mathbb{R}^{D}$ is defined as a vector across channels at position $i$ on the 2D lattice of the feature map. The $j^{th}$ vMF kernel is defined as $\boldsymbol \mu_{j} \in \mathbb{R}^{D}$. With Eq.~\ref{eq::vmf}, we can obtain the vMF activations $\mathbf{Z}_{vMF}$.} \label{fig::decomp}
% \end{figure}

\subsubsection{Compositionality}
Following~\cite{stone2017teaching}, we define a compositional representation as satisfying:

\begin{equation}
\label{eq::compositionality}
    \bm{F}_\psi(S\circ \mathbf{X}) = S\circ \bm{F}_\psi(\mathbf{X}),
\end{equation}
where $S\circ$ denotes the separation operation. If the representation of the separated generative factor in $\mathbf{X}$ is equivalent to the separated representation of $\mathbf{X}$ using the same separation operation, then the representation $S\circ\bm{F}_\psi(\mathbf{X})$ is compositional. For example, the separation operation can be masking the image with the masks of objects as in~\cite{stone2017teaching}. Typically, designing such separation operations requires knowing the ground truth generative factors.

\subsubsection{Compositional equivariance}
Equivariance~\cite{liu2022learning} denotes:
\begin{equation}
\label{eq::equivariance}
    \bm{F}_\psi(M_g\cdot \mathbf{X}) = M_g\cdot \bm{F}_\psi(\mathbf{X}),
\end{equation}
where $M_g$ denotes a set of transformations. Here, $\bm{F}_\psi(\mathbf{X})$ is equivariant if there exists $M_g$ such that the transformations of the input $\mathbf{X}$ that transform the output $\bm{F}_\psi(\mathbf{X})$ in the same manner. We then define a \textbf{compositionally equivariant} representation as satisfying:
\begin{equation}
\label{eq::compequivariance}
    \bm{F}_\psi(M_g\cdot S\circ \mathbf{X}) = M_g\cdot S\circ \bm{F}_\psi(\mathbf{X}).
\end{equation}
This implies that a representation is compositionally equivariant if it represents a generative factor that is defined by performing the separation operation on $\mathbf{X}$ and there exist transformations that equivariantly affect the factor in the $\mathcal{X}$ space and in the $\mathcal{Z}$ space. In the real world, the generative factors are usually indeed compositionally equivariant especially when we consider equivariant tasks like segmentation, registration, etc. For example, the heart can be separated out in the cardiac MRI images as in Fig.~\ref{fig::concept}. Performing transformations on the generative factor of the heart (i.e. shrinking the heart anatomy) will equivariantly transform the cardiac MRI image (i.e. representing the shrunk heart anatomy).

% A simple example is that considering car wheels as a generative factor, composing car wheels with other car components (equivalent to performing transformations on the car wheels) can represent different cars, which does not affect the separation of the car wheels from different cars. We claim that when the learnt latent representations satisfy compositional equivariance, the representations approximate well the ground truth generative factors. 

% \todo{Change the example and relate to new Figure 1.}
% \todo[inline]{@Sotos, I feel it is a bit unconvincing that the generative factors must be compositionally equivariant. I think for equivariant tasks, the so-called "object-centric" representations are compositionally equivariant. But for other tasks, the argument that the factors are compositionally equivariant may not hold.}

\subsubsection{Compositionally equivariant representations}
To learn a compositionally equivariant representation, the key is to find a proper separation operation or its approximation and to design the transformations. Motivated by~\cite{stammer2022interactive, locatello2020weakly, higgins2018towards, wang2021selfsupervised}, we assume that it is known that for a group of data samples $\{\mathbf{X_k^1}, \cdots,\mathbf{X_k^{N_k}}\}$, there exists at least one generative factor that is shared across all samples. In this case, comparing $\{\mathbf{X_k^1}, \cdots,\mathbf{X_k^{N_k}}\}$, we can identify the shared factor.
If we compose the shared factor with different factors to generate the different data $\{\mathbf{X_k^1}, \cdots,\mathbf{X_k^{N_k}}\}$, this is equivalent to performing transformations on the shared factor. Hence, with the limited information that the data group shares some factors, we can design an objective to train the model to learn compositionally equivariant representations. In particular, for any $i \in {\{1, \cdots, N_k\}}$ and $h\in {\{1, \cdots, N_k\}}$, we aim to minimise the compositionally equivariant objective:

\begin{equation}
\label{eq::compequivarianceloss}
    \mathcal{L}^{i, h} =  |\bm{F}_\psi(\mathbf{X_k^i})_{j} - \bm{F}_\psi(\mathbf{X_k^h})_{j}|_1,
\end{equation}
where $j$ denotes the index of the shared factor. Note that directly minimising Eq.~\ref{eq::compequivarianceloss} requires knowing which factors are shared across the data group, which is a strong assumption, especially for medical data. Hence, it is more feasible to design specific learning objectives or design biases to \textit{implicitly} minimise Eq.~\ref{eq::compequivarianceloss}. In the following, we study several different approaches that implicitly achieve compositional equivariance.

\subsection{Modeling compositional representations}
We first model compositional representations with the learnable von-Mises-Fisher (vMF) kernels as shown in Fig.~\ref{fig::allmodel} top left. In other words, we represent deep features in a compact low dimensional vMF space. We denote the features extracted by $\bm{F}_\psi$ as $\mathbf{Z} \in \mathbb{R}^{H\times W\times D}$, where $H$ and $W$ are the spatial dimensions and $D$ is the number of channels. The feature vector $\mathbf{z}_i\in \mathbb{R}^{D}$ is defined as a vector across channels at position $i$ on the 2D lattice of the feature map. We follow Compositional Networks~\cite{kortylewski2020compositional} to model $\mathbf{Z}$ with $J$ vMF distributions, where the learnable mean of the $j^{th}$ vMF kernel distribution is defined as $\boldsymbol \mu_{j} \in \mathbb{R}^{D}$. To make the modelling tractable, the variance $\sigma$ of all distributions is fixed. In particular, the vMF activation for the $j^{th}$ distribution at each position $i$ can be calculated as: 
% \begin{equation}
%     p(\mathbf{z}_i|\boldsymbol \mu_{j}) = {C(\sigma)}^{-1} \cdot {e^{\sigma \boldsymbol \mu_{j}^{T} \mathbf{z}_i}} , \text{ s.t. } ||\boldsymbol \mu_{j}||=1, 
% \end{equation}
\begin{equation}
   z_{i, j} \equiv p(\mathbf{z}_i|\boldsymbol \mu_{j}) = \frac{e^{\sigma_{j} \boldsymbol \mu_{j}^{T} \mathbf{z}_i}}{C(\sigma)_{j}} , \text{ s.t. } ||\boldsymbol \mu_{j}||=1, 
\label{eq::vmf}
\end{equation}
where $||\mathbf{z}_i||=1$ and $C(\sigma)$ is a constant. After modelling the image features with $J$ vMF distributions according to Eq.~\ref{eq::vmf}, the tensor of vMF activations $\mathbf{Z}_{vMF} \in \mathbb{R}^{H\times W\times J}$ can be obtained, indicating how much each kernel is activated at each position. We leverage the compositional kernels as compositional representations. However, simply decomposing the features into a compositional latent space does not ensure the assignment of \textit{meaningful} information to each compositional representation i.e. achieving compositional equivariance.

% \begin{figure}[t]
% \includegraphics[width=0.5\textwidth]{un.pdf}
% \centering
% \caption{Unsupervised compositionally equivariant representation learning model. We train the vMF kernels with Eq.~\ref{eq::vmfloss}. $\bm{F}_\psi$ is the encoding part of a U-Net that is pre-trained to reconstruct the input image.}
% \label{fig::uncermodel}
% \end{figure}

% \begin{figure*}[ht]
% \includegraphics[width=1\textwidth]{weakly.pdf}
% \centering
% \caption{Overall model design for weakly supervised compositionally equivariant representation learning. The image is first encoded and then the vMF activations are calculated as the input of the classifier. We use the presence or absence of heart in the image as weak supervision.} \label{fig::weakly}
% \end{figure*}

\subsection{Achieving compositional equivariance}
The decomposition process described above allows us to extract compositional representations. However, these representations are not bound to be compositionally equivariant. In other words, the decomposed representations usually do not correspond to the underlying generative factors. We consider three different settings that can assign corresponding generative factors' information to the compositional representations in order to achieve compositional equivariance. 

\subsubsection{Unsupervised setting}

We first consider that no supervision information is provided. We use the clustering loss in~\cite{liu2022vmfnet} to enforce the compositional representations to correspond to the centres of any clusters of the input feature vectors (as in Fig.~\ref{fig::allmodel} top left). The loss $\mathcal{L}_{clu}$ that forces the kernels to be the cluster centres of the feature vectors is defined in~\cite{kortylewski2020compositional} as:

\begin{equation}
    \mathcal{L}_{clu}(\boldsymbol \mu, \mathbf{Z}) = -{(HW)}^{-1}\sum_{i} \operatorname*{max}_{j} \boldsymbol \mu_{j}^T \mathbf{z}_i,   
\label{eq::vmfloss}
\end{equation}
where we only train the kernels and the feature vectors are fixed and produced by the encoding network $\bm{F}_\psi$.  Note that $\bm{F}_\psi$ is the encoding part of a U-Net that is pre-trained to reconstruct the input image. If the group of data that shares some factors forms a cluster in latent space, then using the clustering loss will possibly align the kernels with the cluster centres of the data groups. One can expect that the assumption of groups of data forming clusters is not always true in practice. Also, multiple kernels may be aligned to the same cluster centre. It is likely to be that with the clustering loss, the compositional representations can capture part of the information of the factors i.e. achieving a certain level of compositional equivariance. 

\subsubsection{Weakly supervised setting}
Next, we consider using weak supervision describing whether or not a given shared factor is present in each image (e.g. \textit{heart} in cardiac images). Note that we consider the task of medical image segmentation in this paper. Hence, to help with downstream tasks, it is important to consider the shared factors that are corresponded to the task. In this case, we can learn compositionally equivariant representations of the heart and potentially use the activations for heart localisation and segmentation. We define the label as $c$ which indicates the presence or absence of the heart in the image. Here, the task network is a binary classifier i.e. $\hat{c} = \bm{T}_{\theta_{C}}(\mathbf{Z}_{vMF})$. The weak supervision loss is:

\begin{equation}
    \mathcal{L}_{weak}(\hat{c}, c) = |\hat{c} - c|_1.
\end{equation}

We combine this weakly supervised loss with the clustering loss to obtain the overall objective:

\begin{equation}
       \operatorname*{argmin}_{\psi, {\theta_{C}}, \boldsymbol \mu} \quad \mathcal{L}_{weak}(\hat{c}, c)  +  \mathcal{L}_{clu}(\boldsymbol \mu, \mathbf{Z}).
\label{eq::weakloss}
\end{equation}

After adding weak supervision about the heart, we expect that some of the learned compositional representations will be assigned corresponding information i.e.compositionally equivariant representations corresponding to the \textit{heart} factor.

\subsubsection{Semi-supervised setting with reconstruction}
\begin{figure}[t]
\includegraphics[width=0.45\textwidth]{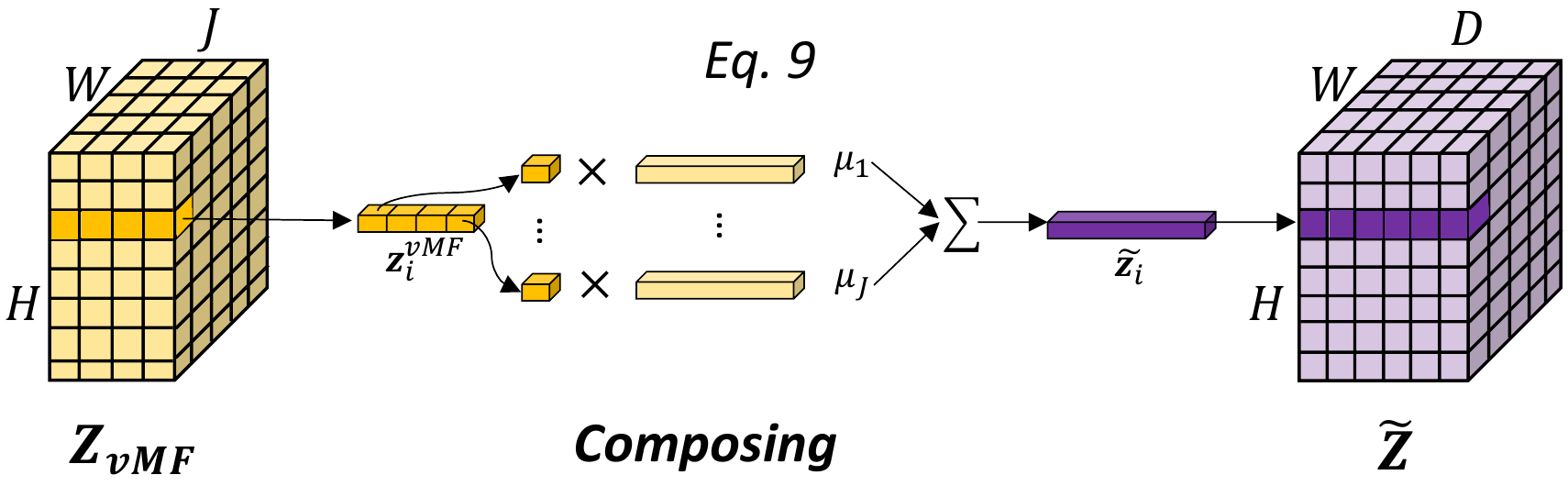}
\centering
\caption{The composing module. We construct a new feature space $\mathbf{\widetilde{Z}}$  (with Eq.~\ref{eq::recompose}) to approximate the encoded features $\mathbf{Z}$, enabling the reconstruction of the input image.} \label{fig::comp}
\end{figure}

We further consider a semi-supervised setting, by leveraging a reconstruction module to train also on data without labels for the downstream segmentation task. As proposed in our previous work (vMFNet)~\cite{liu2022vmfnet}, the model composes the vMF kernels to reconstruct the image with $\bm{R}_\omega$ by using the vMF activations as the composing operations. Then, the vMF activations that contain spatial information are used to predict the segmentation mask with $\bm{T}_\theta$. The composing module is shown in Fig.~\ref{fig::comp}. The overall model design of the vMFNet is shown in Fig.~\ref{fig::allmodel}. Note that using more unlabelled data in training implicitly constructs more groups of data that share the same factors, which enforces implicitly the learnt representation to be more compositionally equivariant.

After decomposing the image features with the vMF kernels and the activations, we re-compose to reconstruct the input image. Reconstruction requires that complete information about the input image is captured~\cite{Achille2017}. In this case, it is possible to observe if the compositional representations have captured information about all the generative factors for the image. However, the vMF activations contain only spatial information, as observed in~\cite{kortylewski2020compositional}, while style information is compressed into the kernels $\boldsymbol \mu_{j}, j\in \{1\cdots J\}$, where the compression is not invertible. Consider that the vMF activation $p(\mathbf{z}_i|\boldsymbol \mu_{j})$ denotes how much the kernel $\boldsymbol \mu_{j}$ is activated by the feature vector $\mathbf{z}_i$. We construct a new feature space $\mathbf{\widetilde{Z}}$ (as in~\cite{liu2022vmfnet}) with the vMF activations and kernels. Let $\mathbf{z}^{vMF}_i\in \mathbb{R}^{J}$ be a normalised vector across $\mathbf{Z}_{vMF}$ channels at position $i$. We devise the new feature vector $\mathbf{\widetilde{z}}_i$ as the combination of the kernels with the normalised vMF activations as the combination coefficients:
\begin{equation}
    \mathbf{\widetilde{z}}_i = \sum_{j=1}^{J} \mathbf{z}^{vMF}_{i, j}  \boldsymbol \mu_{j}, \text{ where } ||\mathbf{z}^{vMF}_{i}||=1.
\label{eq::recompose}
\end{equation}
After obtaining $\mathbf{\widetilde{Z}}$ as the approximation of $\mathbf{Z}$, the reconstruction network $\bm{R}_\omega$ reconstructs the input image with $\mathbf{\widetilde{Z}}$ as the input, i.e. $\mathbf{\hat{X}}=\bm{R}_\omega(\mathbf{\widetilde{Z}})$. The reconstruction loss is defined as:

\begin{equation}
    \mathcal{L}_{rec}(\mathbf{X}, \hat{\mathbf{X}}) = |\mathbf{X} - \hat{\mathbf{X}}|_1,
\end{equation}

As the vMF activations contain only spatial information of the image that is highly correlated to the segmentation mask, we design a segmentation module, i.e. the task network $\bm{T}_\theta$, to predict the segmentation mask with the vMF activations as input, i.e. $\mathbf{\hat Y}=\bm{T}_\theta(\mathbf{Z}_{vMF})$. Specifically, the segmentation mask tells what anatomical part the feature vector $\mathbf{z}_i$ corresponds to, which provides further guidance for the model to learn the vMF kernels as the components of the anatomical parts. Then the vMF activations will be further aligned when trained with multi-domain data and hence perform well on domain generalisation tasks. Overall, the feature vectors of different images corresponding to the same anatomical part will be clustered and activate the same kernels. In other words, the vMF kernels are learnt as the components or patterns of anatomical parts i.e. compositionally equivariant representations. Hence, the vMF activations $\mathbf{Z}_{vMF}$ for the features of different images will be aligned to follow the same distributions (with the same means). In this case, comparing with the content-style disentanglement paradigm~\cite{chartsias2019factorised, liu2020metrics}, the vMF activations can be considered as containing the content information and the vMF kernels as containing the style information.

Overall, the model contains trainable parameters $\psi$, $\theta$, $\omega$ and the kernels $\boldsymbol \mu$. The model can be trained end-to-end with the following objective:
\begin{equation}
\begin{aligned}
    \operatorname*{argmin}_{\psi, \theta, \omega, \boldsymbol \mu} \quad \lambda_{Dice} \mathcal{L}_{Dice}(\mathbf{Y}, \hat{\mathbf{Y}}) + &\\ \mathcal{L}_{rec}(\mathbf{X}, \hat{\mathbf{X}}) + \mathcal{L}_{clu}(\boldsymbol \mu, \mathbf{Z}),
    \label{Eqa::totalloss}
\end{aligned}
\end{equation}
where $\lambda_{Dice}=1$ when the ground truth mask $\mathbf{Y}$ is available, otherwise $\lambda_{Dice}=0$. $\mathcal{L}_{Dice}$ is the Dice loss as defined in~\cite{milletari2016v}. 

\subsubsection{Semi-supervised setting with cross pseudo supervision}
An alternative way to take advantage of unlabelled data for the downstream segmentation task is using cross pseudo supervision as proposed in~\cite{chen2021semi}. In particular, two identical segmentation models that are initialised differently are trained simultaneously, where the pseudo supervision of one model is the output of the other model with the same input. Such cross pseudo supervision is equivalent to ensembling multiple models to minimise the uncertainty of the prediction. Here, we design the segmentation model by directly using the vMF activations as the input to a segmentation module as shown in Fig.~\ref{fig::allmodel}. The cross pseudo supervision (CPS) loss is defined as:

\begin{equation}
    \mathcal{L}_{CPS}(\mathbf{Y}_{pseudo}, \hat{\mathbf{Y}}) = \mathcal{L}_{Dice}(\mathbf{Y}_{pseudo}, \hat{\mathbf{Y}}),
\end{equation}
where $\mathbf{Y}_{pseudo}$ is the pseudo ground truth segmentation mask and is detached during training (to stop gradients). Overall, the model is trained with the following objective:

\begin{equation}
\begin{aligned}
    \operatorname*{argmin}_{\psi, \theta, \boldsymbol \mu, \psi', \theta', \boldsymbol \mu'} \quad \lambda_{Dice} \mathcal{L}_{Dice}(\mathbf{Y}, \hat{\mathbf{Y}}) + \mathcal{L}_{clu}(\boldsymbol \mu, \mathbf{Z'}) + & \\ \lambda_{Dice}\mathcal{L}_{Dice}(\mathbf{Y}, \hat{\mathbf{Y'}}) + \mathcal{L}_{clu}(\boldsymbol \mu' , \mathbf{Z'}) + &\\
    \lambda_{CPS} \mathcal{L}_{CPS}(\hat{\mathbf{Y}}, \hat{\mathbf{Y'}}) + \lambda_{CPS} \mathcal{L}_{CPS}(\hat{\mathbf{Y'}}, \hat{\mathbf{Y}}),
    \label{Eqa::cpstotalloss}
\end{aligned}
\end{equation}
where $\lambda_{Dice}=1$ when the ground truth mask $\mathbf{Y}$ is available, otherwise $\lambda_{Dice}=0$. We set  $\lambda_{CPS}$ as $0.1$. The model is termed vMFPseudo.

\subsubsection{Semi-supervised setting with weak supervision}
For the task of cardiac image segmentation, we can apply the weak supervision of predicting the presence or absence of the left ventricle (LV), myocardium (MYO) and right ventricle (RV). We define the label $\mathbf{c}$ as a three-dimensional vector which indicates the presence or absence of the LV, MYO and RV in the image. We use the output of the segmentation module as the input for the weak supervision classifier ($\theta_{C}$), termed vMFWeak. It is possible to apply weak supervision on the latent space i.e. using the vMF activations as the input for the weak supervision task. However, our early experiments show that this will not help on improving the segmentation performance as for weakly-labelled data (no segmentation masks provided), the segmentation module is not trained. Note that the reconstruction is used similarly as a weak supervision, which takes the vMF activations as the input and indeed helps with segmentation. The difference is that reconstruction ensures that all the information in the images is captured in the latent space, which contains the information for segmentation. However, weak supervision does not require the latent space to capture all the information but as little as much information for the weak supervision tasks, which possibly hurts the segmentation performance. 

Overall, the model contains trainable parameters $\psi$, $\theta_{C}$, $\theta$ and the kernels $\boldsymbol \mu$. The model can be trained with:
\begin{equation}
\begin{aligned}
    \operatorname*{argmin}_{\psi, \theta_{C}, \theta, \boldsymbol \mu} \quad \lambda_{Dice} \mathcal{L}_{Dice}(\mathbf{Y}, \hat{\mathbf{Y}}) +  &\\ \lambda_{weak}\mathcal{L}_{weak}(\hat{\mathbf{c}}, \mathbf{c}) + \mathcal{L}_{clu}(\boldsymbol \mu, \mathbf{Z}),
    \label{Eqa::weaktotalloss}
\end{aligned}
\end{equation}
where $\lambda_{Dice}=1$ when the ground truth mask $\mathbf{Y}$ is available, otherwise $\lambda_{Dice}=0$. $\lambda_{weak}$ is set as $0.5$.

\section{Experiments}

% \begin{figure*}[ht]
% \includegraphics[width=1\textwidth]{cps.pdf}
% \centering
% \caption{Overall model design for semi-supervised compositionally equivariant representation learning with cross pseudo supervision for domain-generalised medical image segmentation. We simultaneously train two models and use the prediction of one model as the pseudo supervision for the other model. The segmentation module is used to predict the segmentation mask by taking the vMF activations as input.} \label{fig::cps}
% \end{figure*}

\begin{figure}[t]
\includegraphics[width=0.4\textwidth]{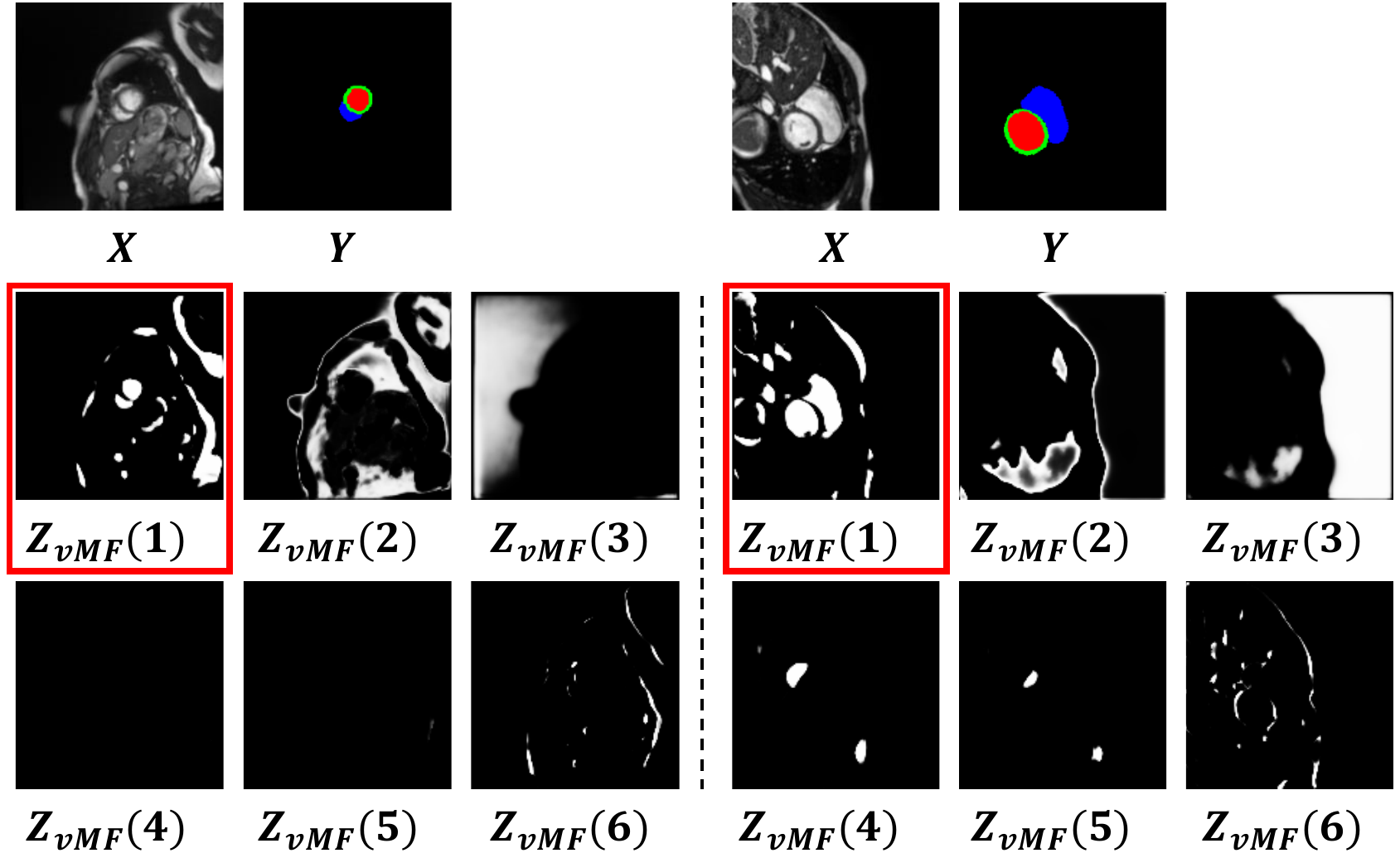}
\centering
\caption{Visualisation of images, ground truth segmentation masks, and 6 out of 12 vMF activation channels for 2 example images using \textbf{the unsupervised setting} from M\&Ms dataset. The channels are manually ordered. The red box highlights the activation of the kernel (partially) corresponding to the heart.}
\label{fig::uncompvisuals}
\end{figure}

\subsection{Datasets}
We adopt the following datasets for our experiments. The \textbf{multi-centre, multi-vendor \& multi-disease cardiac image segmentation (M$\&$Ms) dataset~\cite{mnms}} consists of 320 subjects scanned at 6 clinical centres using 4 different magnetic resonance scanner vendors i.e. domains A, B, C and D. For each subject, only the end-systole and end-diastole phases are annotated. Voxel resolutions range from $0.85\times 0.85\times 10$ mm to $1.45\times 1.45\times 9.9$ mm. Domain A contains 95 subjects, domain B contains 125 subjects, and domains C and D contain 50 subjects each. The \textbf{spinal cord gray matter segmentation (SCGM) dataset~\cite{prados2017spinal}} images are collected from 4 different medical centres with different MRI systems i.e. domains 1, 2, 3 and 4. The voxel resolutions range from $0.25\times0.25\times2.5$ mm to $0.5\times0.5\times5$ mm. Each domain has 10 labelled subjects and 10 unlabelled subjects.

\subsection{Implementation details}
All models are trained using the Adam optimiser~\cite{kingma2014adam} with a learning rate of $1\times e^{-4}$ for 50K iterations using a batch size of 4 for the semi-supervised settings. Images are cropped to $288\times 288$ for M\&Ms and $144\times 144$ for SCGM. $\bm{F}_\psi$ is a 2D U-Net~\cite{ronneberger2015u} without the last upsampling and output layers to extract features $\mathbf{Z}$. Note that $\bm{F}_\psi$ can be replaced by other encoders such as a ResNet~\cite{he2016deep} and the feature vectors can be extracted from any layer of the encoder where performance may vary for different layers. For all settings, we pre-train the U-Net for 50 epochs with unlabelled data from the source domains. For the weakly supervised setting, the classifier $\bm{T}_\theta$ has 5 CONV-BN-LeakyReLU layers (kernel size 4, stride size 2 and padding size 1) and two fully-connected layers that down-sample the features to 16 dimensions and 1 dimension (for output). For the semi-supervised settings, $\bm{T}_\theta$ and $\bm{R}_\omega$ have similar structures, where a double CONV layer (kernel size 3, stride size 1 and padding size 1) in U-Net with batch normalisation and ReLU is first used to process the features. Then a transposed convolutional layer is used to upsample the features followed by a double CONV layer with batch normalisation and ReLU. Finally, an output convolutional layer with $1\times 1$ kernels is used. For $\bm{T}_\theta$, the output of the last layer is processed with a sigmoid operation.

\begin{table*}[t]
\centering
\caption{Average Dice (\%) and Hausdorff Distance (HD) results and the standard deviations on the M\&Ms and SCGM datasets. For semi-supervised approaches, the training data contains all unlabelled data and different percentages of labelled data from source domains. The other approaches are trained with different percentages of the labelled data only. Results of baseline models are taken from~\cite{liu2022vmfnet}. Bold numbers denote the best performance. }\label{tab1}
\begin{tabular}{|c|c|c|c|c|c|c|c|c|c|}
\hline
\textbf{Percent} & \textbf{metrics}\         & \textbf{nnU-Net}   & \textbf{SDNet+Aug.}  & \textbf{LDDG}    & \textbf{SAML}  & \textbf{DGNet} & \textbf{vMFWeak} & \textbf{vMFPseudo}  & \textbf{vMFNet} \\ \cline{1-10}
\multirow{2}{*}{M\&Ms 2\%}     & Dice ($\uparrow$)      & $65.94_{ 8.3}$             & $68.28_{ 8.6}$     & $63.16_{ 5.4}$             & $64.57_{ 8.5}$    & $72.85_{ 4.3}$ & $75.67_{ 5.4}$ & $77.97_{4.7}$ & $\mathbf{78.43_{ 3.6}}$ \\ \cline{2-10}
     & HD ($\downarrow$)     & $20.96_{ 4.0}$             & $20.17_{ 3.3}$     & $22.02_{ 3.5}$             & $21.22_{ 4.1}$    & $19.32_{ 2.8}$ & $17.24_{ 1.9}$  & $16.61_{1.8}$ & $\mathbf{16.56_{ 1.7}}$ \\ \cline{1-10}
\hline
\multirow{2}{*}{M\&Ms 5\%}     & Dice ($\uparrow$)      & $76.09_{ 6.3}$             & $77.47_{ 3.9}$    & $71.29_{ 3.6}$            & $74.88_{ 4.6}$   & $79.75_{ 4.4}$ & $81.43_{ 3.0}$  & $\mathbf{82.55_{2.6}}$ & $82.12_{ 3.1}$ \\ \cline{2-10}
                         & HD ($\downarrow$)      & $18.22_{ 3.0}$             & $18.62_{ 3.1}$    & $19.21_{ 3.0}$            & $18.49_{ 2.9}$   & $17.98_{ 3.2}$ & $15.44_{ 1.5}$  & $\mathbf{15.10_{1.5}}$  & $15.30_{ 1.8}$ \\ \cline{1-10}
\multirow{2}{*}{M\&Ms 100\%}     & Dice ($\uparrow$)      & $84.87_{ 2.5}$            & $84.29_{ 1.6}$    & $85.38_{ 1.6}$            & $83.49_{ 1.3}$   & $\mathbf{86.03_{ 1.7}}$ & $85.59_{ 1.9}$ & $85.49_{1.6}$ & $85.92_{ 2.0}$ \\ \cline{2-10}
                         & HD ($\downarrow$)      & $14.80_{ 1.9}$            & $15.06_{ 1.6}$    & $14.88_{ 1.7}$            & $15.52_{ 1.5}$   & $14.53_{ 1.8}$ & $13.98_{ 1.1}$  & $\mathbf{13.99_{1.1}}$ & $14.05_{ 1.3}$ \\ \cline{1-10}
\hline
\multirow{2}{*}{SCGM 20\%}     & Dice ($\uparrow$)       & $64.85_{ 5.2}$    & $76.73_{ 11}$           & $63.31_{ 17}$   & $73.50_{ 12}$               & $79.58_{ 11}$ & - & $75.58_{11}$ & $\mathbf{81.11_{ 8.8}}$ \\ \cline{2-10}
     & HD ($\downarrow$)      & $3.49_{ 0.49}$    & $2.07_{ 0.36}$           & $2.38_{ 0.39}$   & $2.11_{ 0.37}$               & $1.97_{ 0.30}$  & - & $2.17_{0.36}$ & $\mathbf{1.96_{ 0.31}}$ \\ \cline{2-10}
\hline
\multirow{2}{*}{SCGM 100\%}     & Dice ($\uparrow$)       & $71.51_{ 5.4}$    & $81.37_{ 11}$           & $79.29_{ 13}$   & $80.95_{ 13}$              & $82.25_{ 11}$ & - & $\mathbf{85.01_{5.8}}$ & $84.03_{ 8.0}$ \\ \cline{2-10}
                         & HD ($\downarrow$)      & $3.53_{ 0.45}$    & $1.93_{ 0.36}$           & $2.11_{ 0.41}$   & $1.95_{ 0.38}$              & $1.92_{ 0.31}$ & - & $1.89_{0.25}$ & $\mathbf{1.84_{ 0.31}}$ \\ \cline{2-10}
\hline
\end{tabular}
\end{table*}

\begin{table*}[ht]
\centering
\caption{Dice (\%) results and the standard deviations on M\&Ms dataset. Bold numbers denote the best performance.}\label{tabA1_vmfnet}
% \resizebox{1\textwidth}{!}{%
\begin{tabular}{|c|c|c|c|c|c|c|c|c|c|c|}
\hline
\multicolumn{2}{|c|}{\textbf{Source}}     & \textbf{Target}           & \textbf{nnU-Net}   & \textbf{SDNet+Aug.}  & \textbf{LDDG}    & \textbf{SAML}  & \textbf{DGNet} & \textbf{vMFWeak} & \textbf{vMFPseudo} & \textbf{vMFNet}  \\ \cline{1-11}
\multirow{ 5}{*}{2\%} & B,C,D  & A                  & $52.87_{ 19}$             & $54.48_{ 18}$     & $59.47_{ 12}$             & $56.31_{ 13}$    & $66.01_{ 12}$ & $66.54_{ 17}$ & $70.12_{16}$ & $\mathbf{73.13_{ 9.6}}$ \\ \cline{2-11}

& A,C,D       & B                                   & $64.63_{ 17}$             & $67.81_{ 14}$     & $56.16_{ 14}$             & $56.32_{ 15}$    & $72.72_{ 10}$ & $77.34_{ 11}$ & $\mathbf{78.77_{10}}$ & $77.01_{ 7.9}$ \\ \cline{2-11}

& A,B,D       & C                                   & $72.97_{ 14}$             & $76.46_{ 12}$     & $68.21_{ 11}$             & $75.70_{ 8.7}$   & $77.54_{ 10}$ & $80.75_{9.4}$ & $\mathbf{81.75_{8.6}}$ & $81.57_{ 8.1}$ \\ \cline{2-11}

& A,B,C       & D                                   & $73.27_{ 11}$             & $74.35_{ 11}$     & $68.56_{ 10}$             & $69.94_{ 9.8}$   & $75.14_{ 8.4}$ & $78.03_{9.8}$ & $81.23_{7.0}$ & $\mathbf{82.02_{ 6.5}}$ \\ \cline{2-11}
\hline
\multirow{ 5}{*}{5\%} & B,C,D   & A                 & $65.30_{ 17}$             & $71.21_{ 13}$     & $66.22_{ 9.1}$            & $67.11_{ 10}$    & $72.40_{ 12}$ & $76.41_{ 8.4}$ & $\mathbf{78.06_{8.8}}$  & $77.06_{ 10}$\\ \cline{2-11}

& A,C,D       & B                                   & $79.73_{ 10}$             & $77.31_{ 10}$     & $69.49_{ 8.3}$            & $76.35_{ 7.9}$   & $80.30_{ 9.1}$ & $\mathbf{83.74_{ 6.7}}$ & $83.49_{7.1}$ & $82.29_{ 7.8}$ \\ \cline{2-11}

& A,B,D       & C                                   & $78.06_{ 11}$             & $81.40_{ 8.0}$    & $73.40_{ 9.8}$            & $77.43_{ 8.3}$   & $82.51_{ 6.6}$ & $81.91_{ 7.5}$ & $83.71_{7.3}$  & $\mathbf{84.01_{ 7.3}}$ \\ \cline{2-11}    

& A,B,C       & D                                   & $81.25_{ 8.3}$            & $79.95_{ 7.8}$    & $75.66_{ 8.5}$            & $78.64_{ 5.8}$   & $83.77_{ 5.1}$ & $83.65_{ 5.6}$ & $84.93_{6.1}$ & $\mathbf{85.13_{ 6.1}}$ \\ \cline{2-11}
\hline
\multirow{ 5}{*}{100\%} & B,C,D  & A                & $80.84_{ 11}$             & $81.50_{ 7.7}$    & $82.62_{ 6.3}$            & $81.33_{ 7.2}$   & $\mathbf{83.21_{ 7.4}}$ & $82.46_{6.7}$ & $82.72_{7.1}$ & $82.67_{ 7.2}$ \\ \cline{2-11}

& A,C,D       & B                                   & $\mathbf{86.76_{ 5.8}}$   & $85.04_{ 6.1}$    & $85.68_{ 5.7}$            & $84.15_{ 5.9}$   & $86.53_{ 5.3}$ & $86.07_{5.3}$ & $86.56_{4.9}$ & $85.95_{ 5.6} $ \\ \cline{2-11}

& A,B,D       & C                                   & $84.92_{ 7.1}$            & $85.64_{ 6.5}$    & $86.49_{ 6.3}$            & $84.52_{ 6.2}$   & $87.22_{ 6.1}$ & $86.33_{5.9}$ & $85.86_{7.5}$ & $\mathbf{87.80_{ 4.4}}$  \\ \cline{2-11}

& A,B,C       & D                                   & $86.94_{ 5.9}$            & $84.96_{ 5.2}$    & $86.73_{ 6.1}$            & $83.96_{ 5.9}$   & $87.16_{ 4.9}$ & $\mathbf{87.49_{4.9}}$ & $86.81_{4.5}$  & $87.26_{ 4.7}$\\ \cline{2-11}
\hline
\end{tabular}
% }
\end{table*}

\begin{table*}[ht]
\centering
\caption{Hausdorff Distance results and the standard deviations on M\&Ms dataset. Bold numbers denote the best performance.}\label{tabA1_hsd_vmfnet}
% \resizebox{1\textwidth}{!}{%
\begin{tabular}{|c|c|c|c|c|c|c|c|c|c|c|}
\hline
\multicolumn{2}{|c|}{\textbf{Source}}     & \textbf{Target}           & \textbf{nnU-Net}   & \textbf{SDNet+Aug.}  & \textbf{LDDG}    & \textbf{SAML}  & \textbf{DGNet} & \textbf{vMFWeak} & \textbf{vMFPseudo} & \textbf{vMFNet}  \\ \cline{1-11}
\multirow{ 5}{*}{2\%} & B,C,D  & A                  & $26.48_{ 7.5}$             & $24.69_{ 7.0}$     & $25.56_{ 5.9}$             & $25.57_{ 5.7}$    & $23.55_{ 6.5}$ & $20.22_{ 6.5}$ & $19.51_{6.2}$ & $\mathbf{19.14_{ 4.8}}$ \\ \cline{2-11}

& A,C,D       & B                                   & $23.11_{ 6.8}$             & $21.84_{ 6.2}$     & $25.44_{ 5.2}$             & $24.91_{ 5.5}$    & $19.95_{ 6.3}$ & $17.22_{ 5.2}$ & $\mathbf{16.84_{5.3}}$ & $17.01_{ 3.7}$ \\ \cline{2-11}

& A,B,D       & C                                   & $16.75_{ 4.6}$             & $16.57_{ 4.2}$     & $18.98_{ 3.9}$             & $16.46_{ 3.5}$   & $16.29_{ 4.0}$ & $15.12_{3.7}$ & $\mathbf{15.06_{3.7}}$ & $15.30_{ 3.5}$ \\ \cline{2-11}
& A,B,C       & D                                   & $17.51_{ 4.9}$             & $17.57_{ 4.1}$     & $18.08_{ 3.8}$             & $17.94_{ 3.8}$   & $17.48_{ 4.7}$ & $16.38_{4.3}$ & $15.04_{3.2}$  & $\mathbf{14.80_{ 3.0}}$\\ \cline{2-11}
\hline
\multirow{ 5}{*}{5\%} & B,C,D   & A                 & $23.04_{ 6.7}$             & $22.84_{ 6.3}$     & $23.35_{ 5.7}$            & $23.10_{ 5.9}$    & $22.55_{ 6.6}$ & $17.91_{ 4.9}$ & $\mathbf{17.54_{4.9}}$ & $18.19_{ 4.9}$ \\ \cline{2-11}

& A,C,D       & B                                   & $18.18_{ 4.7}$             & $20.26_{ 5.5}$     & $20.56_{ 4.7}$            & $18.97_{ 4.9}$   & $19.37_{ 6.4}$ & $14.97_{ 3.9}$ & $\mathbf{14.86_{4.2}}$  & $15.24_{ 3.2}$\\ \cline{2-11}

& A,B,D       & C                                   & $16.44_{ 4.2}$             & $16.22_{ 3.9}$    & $17.14_{ 3.3}$            & $16.29_{ 3.2}$   & $15.77_{ 3.8}$ & $14.91_{ 3.2}$ & $14.35_{3.3}$  & $\mathbf{14.17_{ 3.3}}$\\ \cline{2-11}

& A,B,C       & D                                   & $15.24_{ 4.2}$            & $15.15_{ 3.3}$    & $15.80_{ 3.2}$            & $15.58_{ 3.2}$   & $14.24_{ 2.8}$ & $13.96_{ 2.9}$ & $13.64_{2.8}$  & $\mathbf{13.61_{ 2.8}}$\\ \cline{2-11}
\hline
\multirow{ 5}{*}{100\%} & B,C,D  & A                & $17.86_{ 5.5}$             & $17.39_{ 4.5}$    & $17.48_{ 4.1}$            & $17.70_{ 4.2}$   & $17.28_{ 3.9}$ & $\mathbf{15.80_{3.9}}$ & $15.82_{3.9}$  & $15.99_{ 3.5}$\\ \cline{2-11}

& A,C,D       & B                                   & $14.82_{ 3.4}$   & $15.55_{ 3.7}$    & $15.42_{ 3.4}$            & $16.05_{ 3.7}$   & $14.99_{ 3.6}$ & $13.96_{3.2}$ & $\mathbf{13.94_{3.2}}$  & $14.58_{ 3.2}$\\ \cline{2-11}

& A,B,D       & C                                   & $13.72_{ 3.3}$            & $13.67_{ 3.0}$    & $13.52_{ 2.8}$            & $14.21_{ 3.3}$   & $13.11_{ 2.8}$ & $13.16_{2.8}$ & $13.12_{3.1}$  & $\mathbf{12.70_{ 2.8}}$\\ \cline{2-11}

& A,B,C       & D                                   & $12.81_{ 3.4}$            & $13.64_{ 2.9}$    & $13.11_{ 3.0}$            & $14.12_{ 2.8}$   & $\mathbf{12.72_{ 2.6}}$  & $13.00_{2.6}$ & $13.07_{2.5}$  & $12.94_{ 2.5}$\\  \cline{2-11}
\hline
\end{tabular}
% }
\end{table*}

We follow~\cite{kortylewski2020compositional} to set the variance of the vMF distributions to 30. The number of kernels is set to 12, as it was found empirically in early experiments that this number performed the best. For different medical datasets, the best number of kernels may be slightly different. All models are implemented in PyTorch~\cite{paszke2019pytorch} and are trained using an NVIDIA 2080 Ti GPU. In semi-supervised settings, we use specific percentages of the subjects as labelled data and the rest as unlabelled data. We train the models with 3 source domains and treat the 4$^{th}$ domain as the target one. We use Dice (expressed as \%)~\cite{dice1945measures} and Hausdorff Distance (HD)~\cite{dubuisson1994modified} as the evaluation metrics.

\begin{figure}[t]
\includegraphics[width=0.4\textwidth]{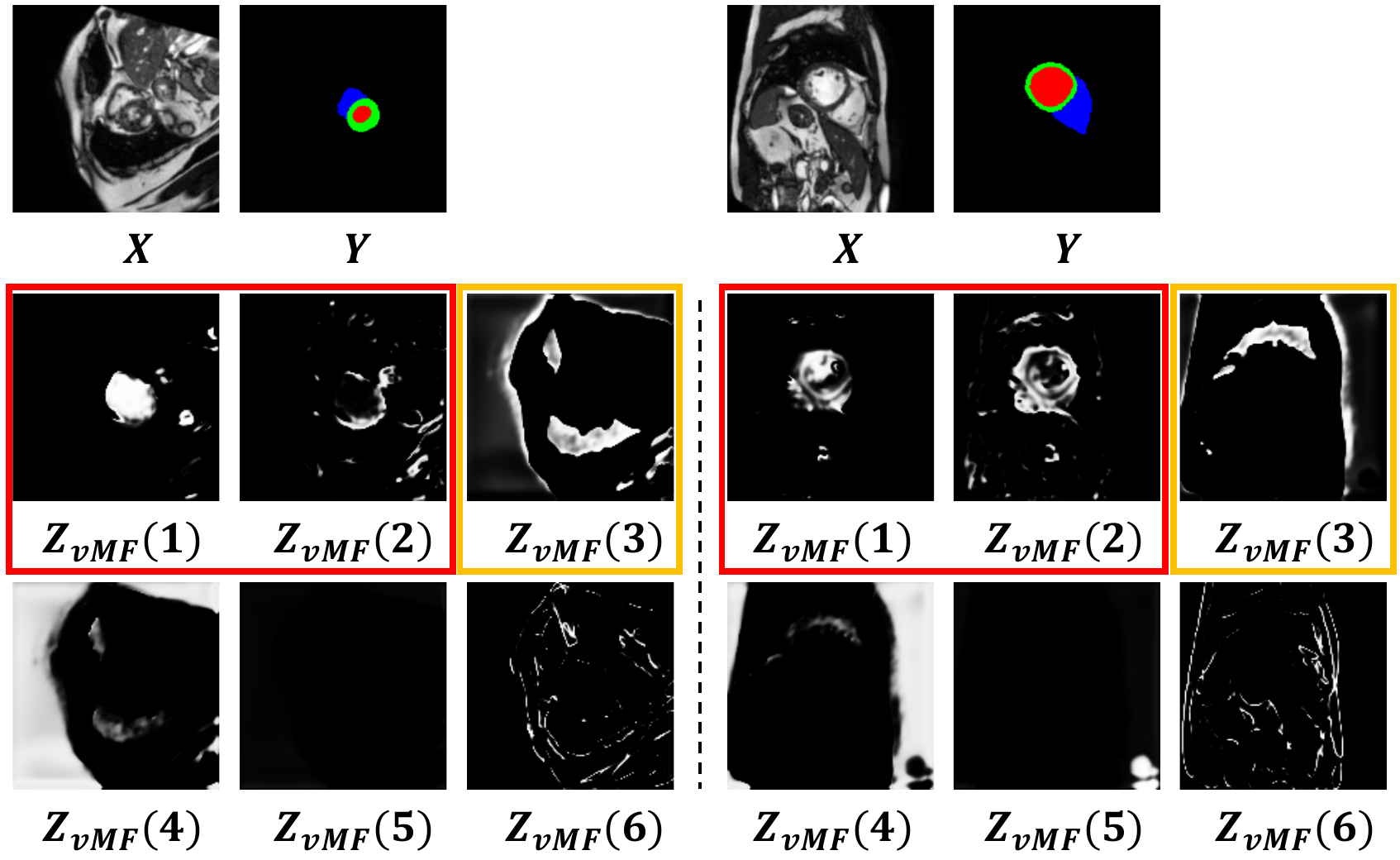}
\centering
\caption{Visualisation of images, ground truth segmentation masks, and 6 out of 12 vMF activation channels for 2 examples of \textbf{the weakly supervised setting} from M\&Ms dataset. The channels are manually ordered. The red box highlights the activation of the kernel (partially) corresponding to the heart. The yellow box relates to the channel that contains information about the lungs.} \label{fig::weakcompvisuals}
\end{figure}

\subsection{Evaluating compositional equivariance}
The generative factors are generalisable and human-understandable. We hence consider how interpretable the activations of the compositionally equivariant representations are and how generalisable the representations are. For interpretability, we follow~\cite{liu2020metrics} to consider how much each vMF activation channel is meaningful (carries information that is relevant to specific anatomy) and how homologous each channel is. For generalisation ability, we consider the performance of the model on the task of semi-supervised domain generalisation as in~\cite{liu2022vmfnet}.  

\subsection{Unsupervised setting}
We train the model as shown in Fig.~\ref{fig::allmodel} top left with Eq.~\ref{eq::vmfloss} for 200 epochs with all the labelled data of the M$\&$Ms dataset. We show the qualitative results in Fig.~\ref{fig::uncompvisuals}. With only the clustering loss, some channels are already meaningful i.e. corresponding to specific anatomy. For example, channel 1 (red box) contains information on the left ventricle (LV) and right ventricle (RV) of the heart. Part of channel 2 is relevant to the lungs. Channel 3 corresponds to the background.

\subsection{Weakly supervised setting}
For the weakly supervised setting, we train the model with Eq.~\ref{eq::weakloss} for 200 epochs with all the labelled data of M$\&$Ms dataset. The qualitative results are shown in Fig.~\ref{fig::weakcompvisuals}. It is clearly shown that a stronger compositional equivariance is achieved compared to the unsupervised setting. Channels 1 and 2 (red box) are more related to the heart. Channel 3 (yellow box) shows the shape of the lungs. Channel 4 contains mostly the background. Overall, the activations of the compositional representations are more interpretable and each channel is more homologous i.e. more compositionally equivariant. Interestingly, for both unsupervised and weakly supervised settings, we observe that one compositional representation represents the lungs even though no information about the lungs is provided. This means that the learnt representations are ready to be used for lung localisation/segmentation when there is a small amount of relevant labelled data available. 

\begin{table*}[ht]
\centering
\caption{Dice (\%) results and the standard deviations on SCGM dataset. Bold numbers denote the best performance.}\label{tabA2}
% \resizebox{1\textwidth}{!}{%
\begin{tabular}{|c|c|c|c|c|c|c|c|c|c|}
\hline
\multicolumn{2}{|c|}{\textbf{Source}}     & \textbf{Target}           & \textbf{nnU-Net}               & \textbf{SDNet+Aug.}                & \textbf{LDDG}                & \textbf{SAML}                           & \textbf{DGNet}  & \textbf{vMFPseudo} & \textbf{vMFNet} \\ \cline{1-10}
\multirow{ 5}{*}{20\%} & 2,3,4  & 1                 & $59.07_{ 21}$        & $83.07_{ 16}$               & $77.71_{ 9.1}$      & $78.71_{ 25}$                  & $87.45_{ 6.3}$  & $87.64_{8.8}$  & $\mathbf{88.08_{ 6.9}}$\\ \cline{2-10}

& 1,3,4       & 2                                   & $69.94_{ 12}$        & $80.01_{ 5.2}$              & $44.08_{ 12}$       & $75.58_{ 12}$                  & $81.05_{ 5.2}$  & $63.50_{16}$ & $\mathbf{81.21_{ 4.2}}$ \\ \cline{2-10}

& 1,2,4       & 3                                   & $60.25_{ 7.2}$       & $58.57_{ 10}$               & $48.04_{ 5.5}$      & $54.36_{ 7.6}$                 & $61.85_{ 7.3}$  & $64.84_{9.3}$  & $\mathbf{66.74_{ 4.9}}$\\ \cline{2-10}    

& 1,2,3       & 4                                   & $70.13_{ 4.3}$       & $85.27_{ 2.2}$              & $83.42_{ 2.7}$      & $85.36_{ 2.8}$                 & $87.96_{ 2.1}$  & $86.35_{2.8}$ & $\mathbf{88.39_{ 2.4}}$ \\ \cline{2-10}
\hline
\multirow{ 5}{*}{100\%} & 2,3,4  & 1                & $75.27_{ 8.3}$    & $90.25_{ 4.5}$  & $88.21_{ 4.9}$   & $90.22_{ 5.6}$              & $90.01_{ 4.9}$  & $89.78_{4.7}$  & $\mathbf{90.96_{ 4.7}}$\\ \cline{2-10}

& 1,3,4       & 2                                   & $76.32_{ 2.9}$    & $84.13_{ 4.2}$           & $83.76_{ 3.1}$   & $\mathbf{86.65_{ 3.5}}$     & $85.48_{ 2.3}$  & $83.39_{4.8}$ & $84.89_{ 3.2}$ \\ \cline{2-10}

& 1,2,4       & 3                                   & $62.59_{ 6.9}$    & $62.18_{ 10}$            & $56.11_{ 9.3}$   & $58.27_{ 9.4}$              & $64.23_{ 9.7}$   & $\mathbf{76.27_{3.7}}$  & $70.71_{ 9.2}$\\ \cline{2-10}

& 1,2,3       & 4                                   & $71.87_{ 2.5}$    & $88.93_{ 1.9}$           & $89.08_{ 2.7}$   & $88.66_{ 2.6}$              & $89.26_{ 2.5}$  & $\mathbf{90.60_{2.0}}$ & $89.57_{ 3.1}$ \\ \cline{2-10}
\hline
\end{tabular}
% }
\end{table*}

\begin{table*}[ht]
\centering
\caption{Hausdorff Distance results and the standard deviations on SCGM dataset. Bold numbers denote the best performance.}\label{tabA2_hsd}
% \resizebox{1\textwidth}{!}{%
\begin{tabular}{|c|c|c|c|c|c|c|c|c|c|}
\hline
\multicolumn{2}{|c|}{\textbf{Source}}     & \textbf{Target}           & \textbf{nnU-Net}               & \textbf{SDNet+Aug.}                & \textbf{LDDG}                & \textbf{SAML}                           & \textbf{DGNet} & \textbf{vMFPseudo} & \textbf{vMFNet}  \\ \cline{1-10}
\multirow{ 5}{*}{20\%} & 2,3,4  & 1                 & $3.09_{ 0.25}$        & $1.52_{ 0.33}$               & $1.75_{ 0.26}$      & $1.53_{ 0.38}$                  & $1.50_{ 0.30}$ & $1.55_{0.34}$ & $\mathbf{1.47_{ 0.33}}$ \\ \cline{2-10}

& 1,3,4       & 2                                   & $3.16_{ 0.09}$        & $1.97_{ 0.16}$              & $2.73_{ 0.33}$       & $2.07_{ 0.35}$                  & $\mathbf{1.91_{ 0.16}}$ & $2.40_{0.39}$ & $1.92_{ 0.14}$  \\ \cline{2-10}

& 1,2,4       & 3                                   & $3.38_{ 0.27}$       & $2.45_{ 0.27}$               & $2.67_{ 0.25}$      & $2.52_{ 0.24}$                 & $\mathbf{2.23_{ 0.23}}$ & $2.43_{0.31}$ & $2.25_{ 0.16}$  \\ \cline{2-10}    

& 1,2,3       & 4                                   & $4.31_{ 0.14}$       & $2.34_{ 0.21}$              & $2.37_{ 0.14}$      & $2.30_{ 0.18}$                 & $2.22_{ 0.13}$ & $2.30_{0.19}$ & $\mathbf{2.18_{ 0.14}}$  \\ \cline{2-10}

\hline
\multirow{ 5}{*}{100\%} & 2,3,4  & 1                & $3.26_{ 0.21}$    & $1.37_{ 0.25}$  & $1.50_{ 0.23}$   & $1.43_{ 0.36}$              & $1.43_{ 0.29}$ & $1.49_{0.32}$ & $\mathbf{1.35_{ 0.25}}$  \\ \cline{2-10}

& 1,3,4       & 2                                   & $3.19_{ 0.09}$    & $1.88_{ 0.16}$           & $2.19_{ 0.19}$   & $\mathbf{1.80_{ 0.19}}$     & $1.81_{ 0.15}$ & $1.88_{0.17}$ & $\mathbf{1.80_{ 0.19}}$  \\ \cline{2-10}

& 1,2,4       & 3                                   & $3.37_{ 0.27}$    & $2.34_{ 0.24}$            & $2.64_{ 0.28}$   & $2.43_{ 0.33}$              & $2.23_{ 0.32}$ & $\mathbf{2.13_{0.21}}$ & $\mathbf{2.13_{ 0.30}}$  \\ \cline{2-10}

& 1,2,3       & 4                                   & $4.30_{ 0.15}$    & $2.13_{ 0.17}$           & $2.12_{ 0.15}$   & $2.15_{ 0.15}$              & $2.11_{ 0.13}$ & $\mathbf{2.06_{0.17}}$ & $2.07_{ 0.18}$  \\ \cline{2-10}

\hline
\end{tabular}
% }
\end{table*}

\begin{figure}[t]
\includegraphics[width=0.4\textwidth]{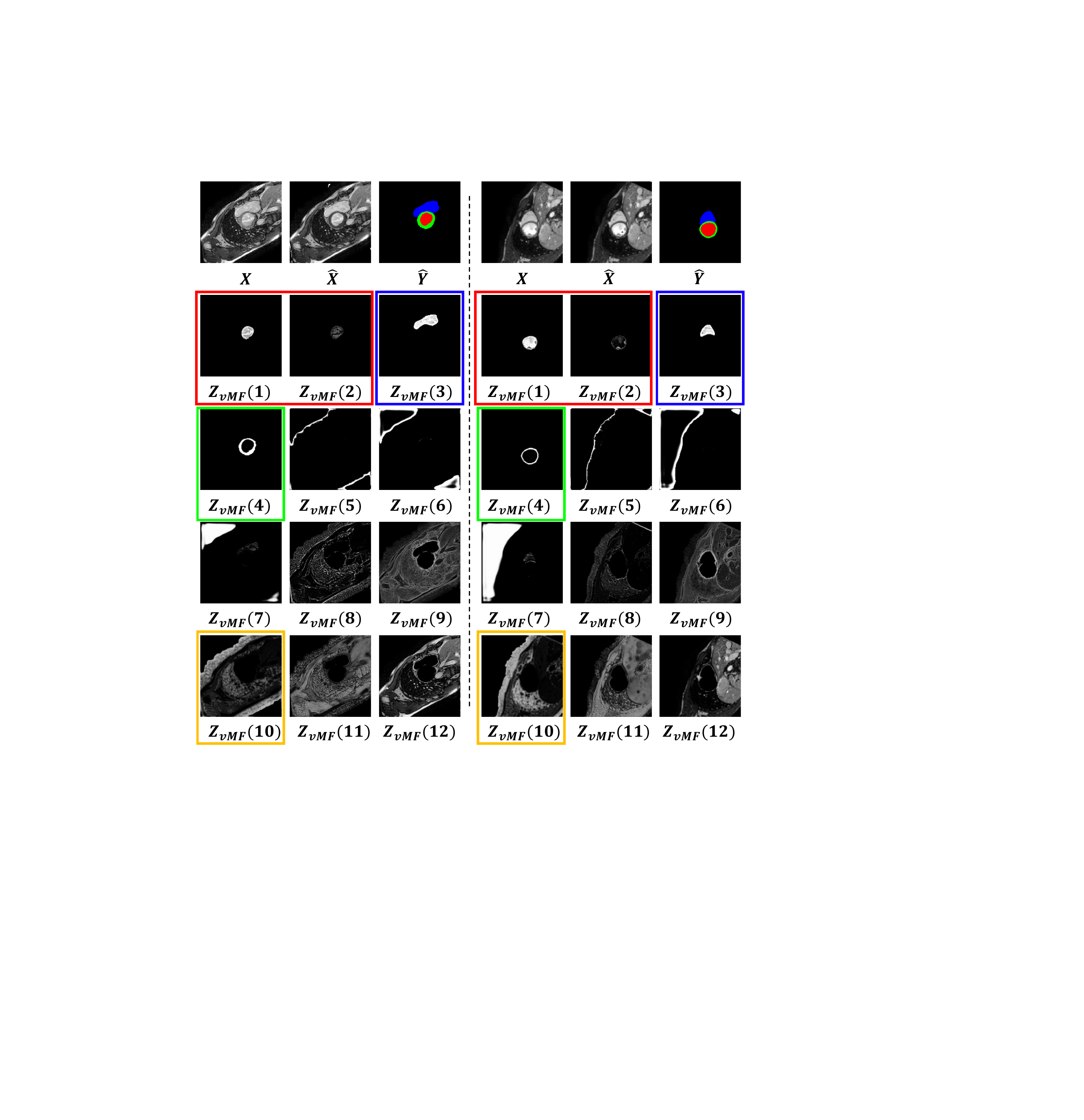}
\centering
\caption{Visualisation of images, reconstructions, predicted segmentation masks and 12 vMF activation channels for 2 examples of \textbf{vMFNet} from M\&Ms dataset. The channels are manually ordered. The red box, blue box and green box highlight the activation of the kernels corresponding to the left ventricle, right ventricle and myocardium. The yellow box relates to the channel of the lungs.} \label{fig::compvisuals}
\end{figure}

\begin{figure}[t]
\includegraphics[width=0.4\textwidth]{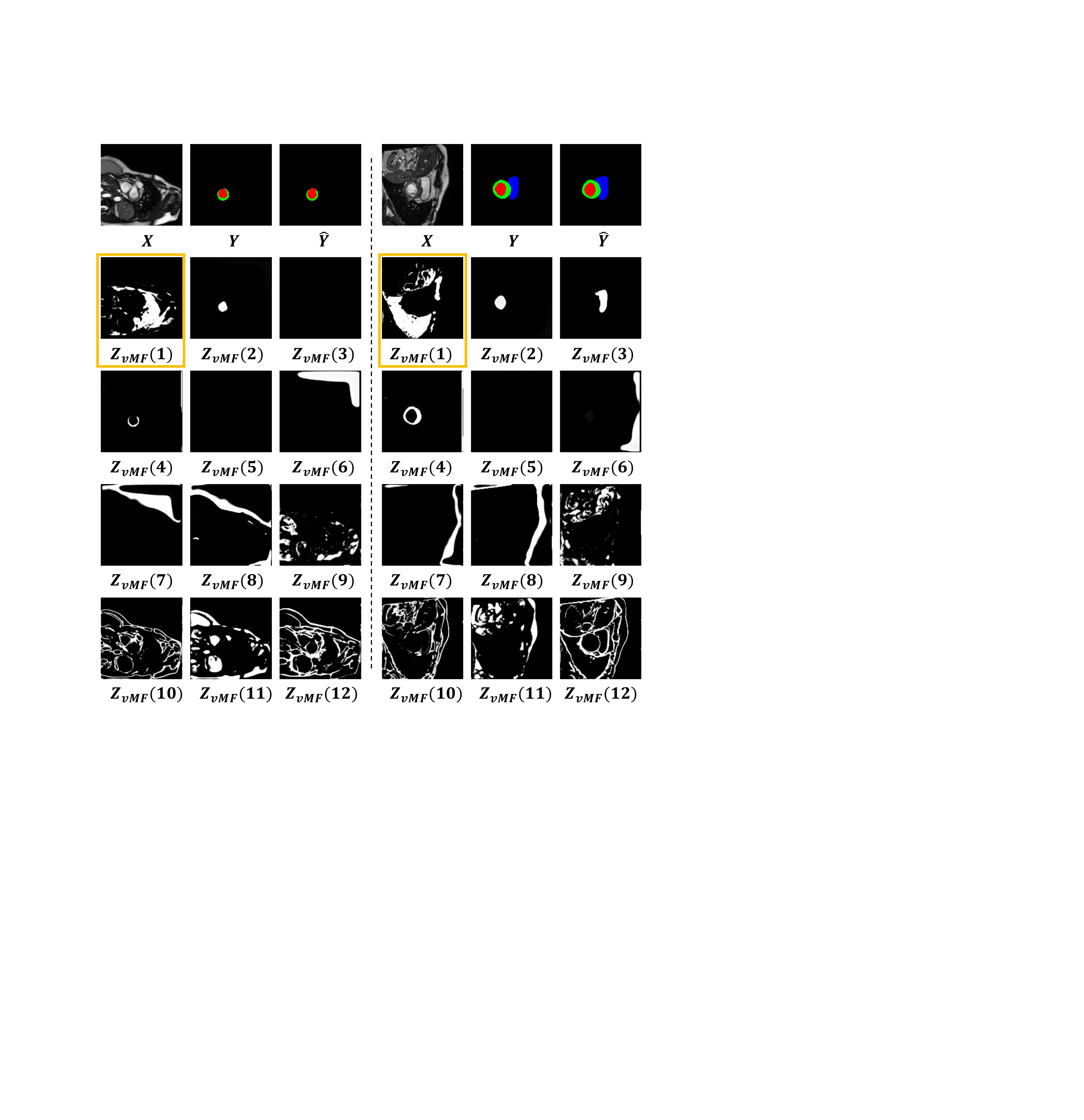}
\centering
\caption{Visualisation of images, ground truth segmentation masks, predicted segmentation masks and 12 vMF activation channels for 2 examples of \textbf{vMFPseudo} from M\&Ms dataset. The channels are manually ordered. The yellow box highlights the channel that contains information about the lungs.} \label{fig::cpsvisuals}
\end{figure}

\subsection{Semi-supervised setting with reconstruction}
For the semi-supervised settings, we test the methods on semi-supervised domain generalisation problems. 
% \begin{figure*}[t]
% \includegraphics[width=\textwidth]{test_rec.png}
% \centering
% \caption{The reconstructed images.} \label{fig::test_rec}
% \end{figure*}

% \begin{figure}[t]
% \includegraphics[width=0.45\textwidth]{traversal.png}
% \centering
% \caption{The traversal images.} \label{fig::traversal}
% \end{figure}

\subsubsection{Baseline models}
For a fair comparison, we compare all models with the same backbone feature extractor, i.e.\ U-Net~\cite{ronneberger2015u}, without any pre-training on other datasets. \textbf{nnU-Net~\cite{isensee2021nnu}} is a supervised baseline. It adapts its model design and searches the optimal hyperparameters to achieve optimal performance. \textbf{SDNet+Aug.~\cite{liu2020disentangled}} is a semi-supervised disentanglement model, which disentangles the input image into spatial anatomy and non-spatial modality factors. Augmenting the training data by mixing the anatomy and modality factors of different source domains, ``SDNet+Aug.'' can potentially generalise to unseen domains. \textbf{LDDG~\cite{li2020domain}} is a fully-supervised domain generalisation model, in which low-rank regularisation is used and the features are aligned to Gaussian distributions. \textbf{SAML~\cite{liu2020shape}} is a gradient-based meta-learning approach. It applies the compactness and smoothness constraints to learn domain-invariant features across meta-train and meta-test sets in a fully supervised setting. \textbf{DGNet~\cite{liu2021semi}} is a semi-supervised gradient-based meta-learning approach. Combining meta-learning and disentanglement, the shifts between domains are captured in the disentangled representations. DGNet achieved the state-of-the-art (SOTA) domain generalisation performance on 
M$\&$Ms and SCGM datasets.

\subsubsection{Generalisation}
\label{sec::results}
Table~\ref{tab1} reports the average results over four leave-one-out experiments that treat each domain in turn as the target domain; more detailed results can be found in Tables~\ref{tabA1_vmfnet} -- \ref{tabA2_hsd}. We highlight that the proposed vMFNet is \textbf{14 times faster to train} compared to the previous SOTA DGNet. Training vMFNet for one epoch takes 7 minutes, while DGNet needs 100 minutes for the M\&Ms dataset due to the expensive meta-test step training in every iteration.

With limited annotations, vMFNet achieves 7.7\% and 3.0\% improvements (in Dice) for 2\% and 5\% cases compared to the previous SOTA DGNet on M\&Ms dataset. For the 100\% case, vMFNet and DGNet have similar performance of around 86\% Dice and 14 HD. Overall, vMFNet has consistently better performance for almost all scenarios on the M\&Ms dataset. Similar improvements are observed for the SCGM dataset. 

\subsubsection{Interpretability}
\label{sec::explain}
Overall, the segmentation prediction can be interpreted as the activation of corresponding compositional representations (kernels) at each position, where false predictions occur when the wrong representations are activated i.e. the wrong vMF activations are used to predict the mask. We show example images, reconstructions, predicted segmentation masks, and the 12 vMF activations channels in Fig.~\ref{fig::compvisuals}. As shown, channels 1 and 2 (red box) are mostly activated by LV feature vectors and channels 3 (blue box) and 4 (green box) are mostly for RV and myocardium (MYO) feature vectors. Interestingly, channel 2 is mostly activated by papillary muscles in the left ventricle even though no supervision about the papillary muscles is provided during training. This supports that the model learns the kernels as the compositionally equivariant representations (patterns of papillary muscles, LV, RV and MYO) of the heart. Although part of channel 10 corresponds to the lungs, the other channels (e.g. channels 8-12) contain mixed (not interpretable and homologous) information about the image as the representations have to contain complete information about the image.

\begin{figure}[t]
\includegraphics[width=0.4\textwidth]{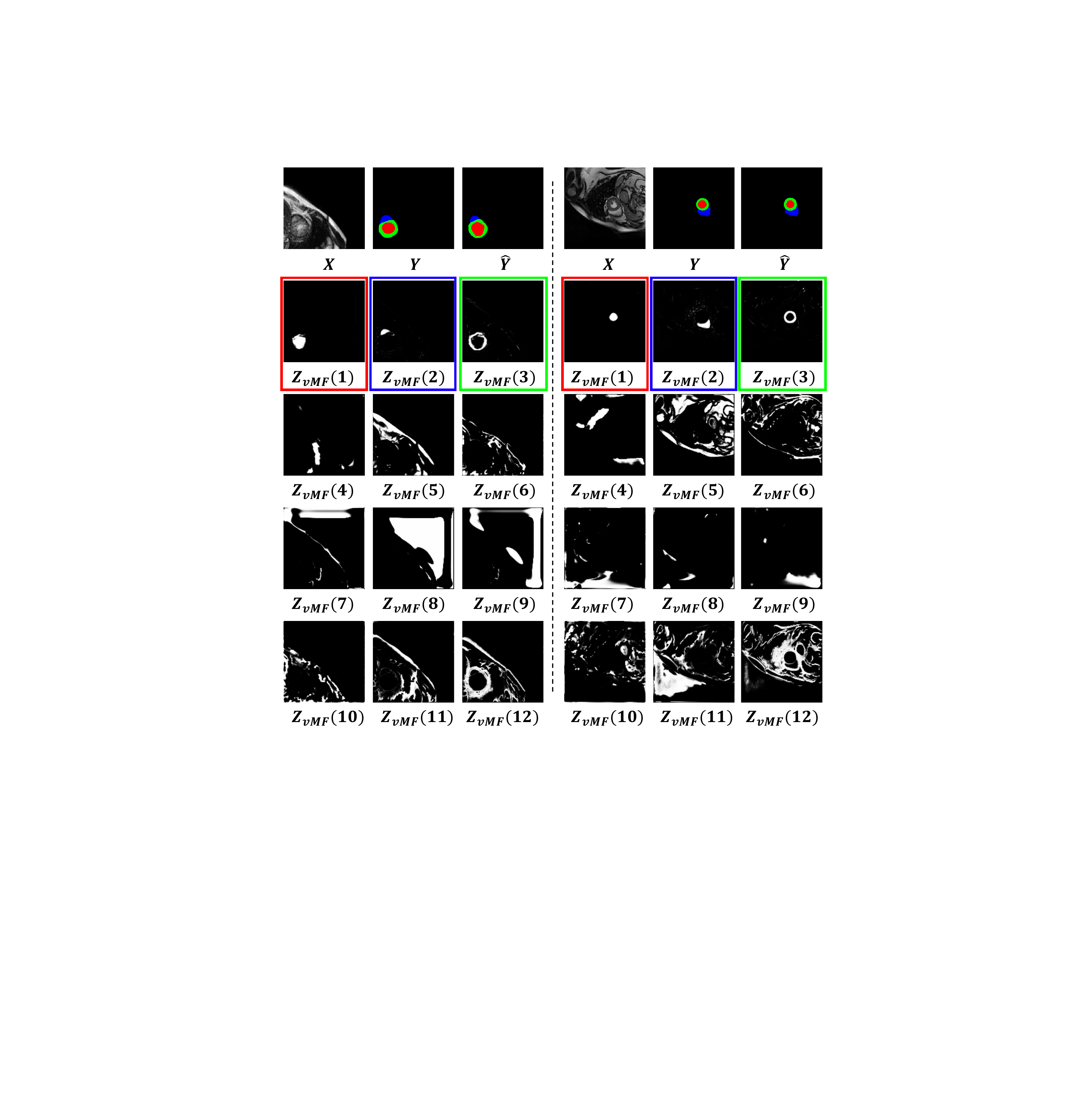}
\centering
\caption{Visualisation of images, reconstructions, predicted segmentation masks and 12 vMF activation channels for 2 examples of \textbf{vMFWeak} from M\&Ms dataset. The channels are manually ordered. The red box, blue box and green box highlight the activation of the kernels corresponding to the left ventricle, right ventricle and myocardium.} \label{fig::semiweakcompvisuals}
\end{figure}

\subsection{Semi-supervised setting with pseudo supervision}
\subsubsection{Generalisation}
The results of vMFPseudo can be found in Table~\ref{tab1}, Table~\ref{tabA1_vmfnet}, Table~\ref{tabA1_hsd_vmfnet}, Table~\ref{tabA2} and Table~\ref{tabA2_hsd}. Notably, vMFPseudo has a similar advantage in the computational load and training speed as vMFNet compared to DGNet. Training vMFPseudo for one epoch takes around 14 minutes, while DGNet needs 100 minutes for the M\&Ms dataset.

Similar to the improvement of vMFNet over the previous SOTA DGNet, vMFPseudo achieves 7.0\% and 3.5\% improvements (in Dice) for 2\% and 5\% cases on the M\&Ms dataset. For the 100\% case, vMFNet is slightly worse than DGNet and vMFNet, which is around 85.5\% Dice and 14 HD. Overall, vMFPseudo consistently performs better for most of the cases compared to the baseline methods for the M\&Ms dataset and SCGM dataset. Compared to vMFNet, we observe that for some cases (e.g. 5\% B,C,D$\rightarrow$A on the M\&Ms dataset and 100\% 1,2,4$\rightarrow$3 on the SCGM dataset), vMFPseudo has clearly better performance. Note that the domain difference between the source domains and the target domain is relatively larger than that in other cases. Hence, the model may produce highly uncertain results for some images in the target domain. In these cases, the cross pseudo supervision loss may help more in mitigating the uncertainty, which produces better results. 

\subsubsection{Interpretability}
Overall, we observe more interpretable results with vMFPseudo as in Fig.~\ref{fig::cpsvisuals}. First of all, the lungs in the images are more clearly shown in channel 1 (yellow box), which means better robustness regarding generalising to other tasks. Channels 2-4 correspond to LV, RV and MYO. As no reconstruction is needed for vMFPseudo, we can see that the other channels are more homologous. For example, channel 10 may relate to the contours of the images.

\subsection{Semi-supervised setting with weak supervision}
For weak supervision, we construct the weak labels for the end-systole and end-diastole phases of the 320 subjects of M\&Ms dataset. Note that the weak supervision does not apply to SCGM data as the gray matter usually exists in every slice.

\subsubsection{Generalisation}
We report the results of vMFWeak in Table~\ref{tab1}, Table~\ref{tabA1_vmfnet}, Table~\ref{tabA1_hsd_vmfnet}. vMFWeak has the same advantage on training speed, where one epoch of training takes around 8 minutes. We observe that vMFWeak similarly outperforms DGNet with 3.9\% and 2.1\% improvements (in Dice) for 2\% and 5\% cases on the M\&Ms dataset. Compared to vMFNet and vMFPesudo, vMFWeak only leverages part of the unlabelled data i.e. the end-systole and end-diastole phases, causing slightly worse performance on the 2\% and 5\% cases. However, for certain cases, vMFWeak still outperforms the other models indicating the effectiveness of the weak supervision.

\subsubsection{Interpretability}
As we show in Fig.~\ref{fig::semiweakcompvisuals}, channels 1-3 correspond to LV, RV and MYO. Due to the constraint of weak supervision, the model is forced to learn a more compact latent space, where most of the information that is irrelevant to the segmentation and weak supervision task is eliminated. Overall, we still can obtain interpretable and homologous representations. However, the representations may not be robust to other tasks.

\section{Conclusion}
In this paper, we have presented that using compositional equivariance as an inductive bias helps to learn generalisable and interpretable compositional representations. In particular, we used different learning biases in different settings to constrain the representations to be compositionally equivariant. For the unsupervised setting and weakly supervised setting, we observed that the representations achieve a certain level of compositional equivalence, which is partially interpretable. For the semi-supervised settings, we qualitatively showed that some of the representations are well interpretable when little supervision is given. Quantitatively, vMFNet, vMFWeak and vMFPseudo, the models built based on decomposing the compositional representations with different design biases and learning biases, achieved the best generalisation performance compared to several strong baselines. Overall, as we discussed in Section~\ref{sec:method} and demonstrated with the results, different learning settings and biases allow the model to learn the representations that are compositionally equivariant at different levels. We conclude that strong prior knowledge (e.g. presence of anatomy) or some supervision significantly boosts the ability to achieve compositional equivariance. Taking advantage of the unlabelled data also plays a key role to learn compositionally equivariant representations as it implicitly constructs more groups of data that have shared factors. 

% For future work on learning compositionally equivariant representations, we suggest constructing weak supervision based on expert knowledge as well as utilising a large scale of unlabelled data with some supervision.

\bibliographystyle{IEEEtran}
\bibliography{main}

\end{document}